\def\year{2018}\relax
\documentclass[letterpaper]{article} 
\usepackage{aaai18}  
\usepackage{times}  
\usepackage{helvet}  
\usepackage{courier}  
\usepackage{url}  
\usepackage{graphicx}  
\usepackage{epsfig}
\usepackage{amsmath}
\usepackage{amssymb}
\usepackage{multirow}
\usepackage{epstopdf}
\usepackage{CJK}
\newcommand{\tabincell}[2]{\begin{tabular}{@{}#1@{}}#2
\end{tabular}}
\begin{document}
%
\title{Cooperative Training of Deep Aggregation Networks for \\RGB-D Action
Recognition}
\author{Pichao Wang$^{\rm 1,2}$, Wanqing Li$^{\rm 1}$, Jun Wan$^{\rm 3}$\thanks{Corresponding author}, Philip Ogunbona$^{\rm 1}$, Xinwang Liu$^{\rm 4}$ \\
$^{\rm 1}$Advanced Multimedia Research Lab, University of Wollongong, Australia\\
$^{\rm 2}$Motovis Inc\\
$^{\rm 3}$Center for Biometrics and Security Research \& National Laboratory of Pattern Recognition\\ Institute of Automation, Chinese Academy of Sciences \\
$^{\rm 4}$School of Computer Science, National University of Defense Technology, Changsha 410073, China \\
{\tt\footnotesize 
\{pw212,wanqing,philipo\}@uow.edu.au,jun.wan@nlpr.ia.ac.cn,xinwangliu@nudt.edu.cn
}\\
}
\maketitle
\begin{abstract}
A novel deep neural network training paradigm that exploits the conjoint information in multiple heterogeneous sources is proposed. Specifically, in a RGB-D based action recognition task, it cooperatively trains a single convolutional neural network (named c-ConvNet) on both RGB visual features and depth features, and deeply aggregates the two kinds of features for action recognition. Differently from the conventional ConvNet that learns the deep separable features for homogeneous modality-based classification with only one softmax loss function, the c-ConvNet enhances the discriminative power of the deeply learned features and weakens the undesired modality discrepancy by jointly optimizing a ranking loss and a softmax loss for both homogeneous and heterogeneous modalities. The ranking loss consists of intra-modality and cross-modality triplet losses, and it reduces both the intra-modality and cross-modality feature variations. Furthermore, the correlations between RGB and depth data are embedded in the c-ConvNet, and can be retrieved by either of the modalities and contribute to the recognition in the case even only one of the modalities is available. The proposed method was extensively evaluated on two large RGB-D action recognition datasets, ChaLearn LAP IsoGD and NTU RGB+D datasets, and one small dataset, SYSU 3D HOI, and achieved state-of-the-art results.
\end{abstract}

\section{Introduction}
RGB-D based action recognition has attracted much attention in recent years due to the advantages that depth information brings to the combined data modality. For example, depth is insensitive to illumination changes and has rich 3D structural information of the scene. However, depth alone is often insufficient for recognizing some actions. In the task of recognizing human-object interactions where texture is vital for successful recognition, depth does not capture the necessary texture context. To exploit the complementary nature of the two modalities, methods~\cite{jia2014latent,Nie2015,Kong2015CVPR,hu2015jointly,wu2015watch,kong2017max} have been proposed to combine the two modalities for RGB-D action recognition and demonstrated the effectiveness of modality fusion. However, most of these methods are based on shallow hand-crafted features and tend to be dataset-dependent. The advent of deep learning has led to the development of methods~\cite{ji20133d,tran2015learning,simonyan2014two,pichao2015,pichaoTHMS,jayaraman2016slow,donahue2015long} based on Convolutional Neural Network (ConvNet) or Recurrent Neural Network (RNN). These methods take as input either RGB or depth or both of them as independent channels and fuse the recognition scores of individual modalities. It is noteworthy that none of these methods address the problem of using heterogeneous inputs (such as RGB and depth) in a cooperative manner to train a single network for action recognition. This cooperative training paradigm allows the powerful representation capability of deep neural network to be fully leveraged and to explore the complementary information in the two modalities using one single network architecture. The need for independent processing channels is thus obviated. Motivated by this observation, in this paper, we propose to adopt deep cooperative neural networks for recognition from the RGB and depth modalities.

One typical challenge in deep learning based action recognition is how a RGB-D sequence could be effectively represented and fed to deep neural networks for recognition. For example, one can conventionally consider it as a sequence of still images (RGB and depth) with some form of temporal smoothness, or as a subspace of images or image features, or as the output of a neural network encoder. Which one among these and other possibilities would result in the best representation in the context of action recognition is not well understood. In addition, it is not clear either how the two heterogeneous RGB and depth channels can be represented and fed into a single deep neural network for the cooperative training.  Inspired by the promising performance of the recently introduced rank pooling machine~\cite{Fernando2015a,bilen2016dynamic} on RGB videos, the rank pooling method is adopted to encode both RGB and depth sequences into compatible dynamic images. A dynamic image contains the temporal information of a video sequence and keeps the spatio-temporal structured relationships of the video; this has been demonstrated to be an effective video descriptor~\cite{bilen2016dynamic}. Based on this pair of dynamic images, namely, RGB visual dynamic images (VDIs) and depth dynamic images (DDIs), a cooperatively trained convolutional neural networks (c-ConvNet) is proposed to exploit the two modality features and enhance the capability of ConvNets for cases in which the features arise from either or both sources.

There are two key issues in using a single c-ConvNet for heterogeneous modalities. First, how to enhance the discriminative power of ConvNets and second, how to reduce the modality discrepancy. Specifically, in most classification cases, the conventional ConvNets can learn separable features but they are often not compact enough to be discriminative~\cite{wen2016discriminative}. Modality discrepancy arises because the feature variations in different modalities pose a challenge for a single network to learn modality-independent features for classification. To handle these two issues, we propose to jointly train a ranking loss  and a softmax loss for action recognition. The ranking loss consists of two intra-modality and cross-modality triplet losses, which reduces variations in both intra-modality and cross-modality. Together with the softmax loss, the signal intra-modality triplet loss enables the c-ConvNet to learn more discriminative features, while the inter-modality triplet loss weakens or eliminates the modalities distribution variations and only focuses on inter-action variations. Moreover, in this way, the correlations between RGB and depth data are embedded in the c-ConvNet, and can be retrieved and contribute to the recognition even in the case where only one of the modalities is available. Furthermore, due to the image structure of dynamic images, the proposed c-ConvNet can be fine-tuned on the pre-trained networks on ImageNet, thus making it possible to work on small datasets. The c-ConvNet was evaluated extensively on three datasets: two large datasets, ChaLearn LAP IsoGD~\cite{wanchalearn} and NTU RGB+D~\cite{shahroudy2016ntu} datasets, and one small one, SYSU 3D HOI~\cite{hu2015jointly} dataset. Experimental results achieved are state-of-the-art.

%

\begin{figure*}[t]
\begin{center}
{\includegraphics[height = 47mm, width = 175mm]{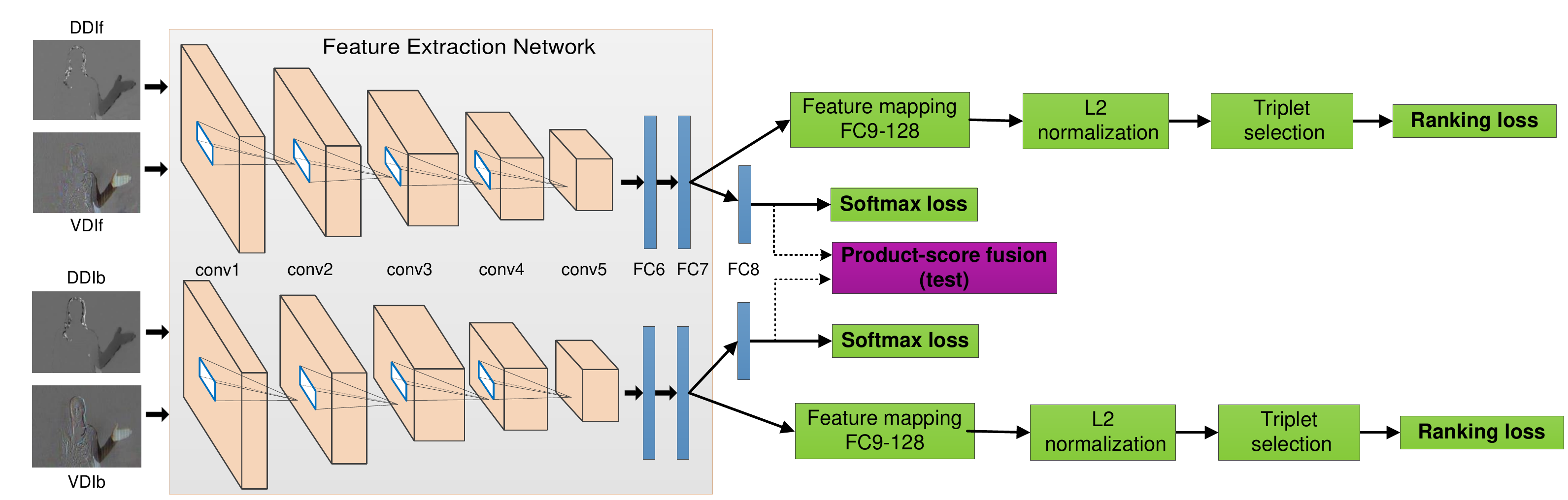}}
\end{center}
\caption{The framework of proposed method. A c-ConvNet consists of one feature
extraction network shared by the ranking loss and softmax loss, and
two separate branches for the two losses. Two distinct c-ConvNets are
adopted to exploit bidirectional information in videos.  The inputs of the two
c-ConvNets are two paired DDIs and VDIs, namely, DDIf \& VDIf, and DDIb \&
VDIb.  During training process, the ranking loss and softmax loss are jointly
optimized; during testing process, an effective product-score fusion method is
adopted for action recognition. The softmax loss serves to learn separable
features for action recognition while the ranking loss encourages the c-ConvNet
to learn discriminative and modality-independent features.}
\label{framework}
\end{figure*}

\section{Related Works}\label{relatedwork}
This paper only reviews the most related depth+RGB fusion-based methods, and for other methods, readers are referred to the survey papers~\cite{aggarwal2014human,presti20163d,zhang2016rgb}. Ni et al.~\cite{ni2011colour} constructed one color-depth video dataset and developed two color-depth fusion techniques based on hand-designed features for human action recognition.  Liu and Shao~\cite{liu2013learning} proposed to adopt genetic programming method to simultaneously extract and fuse the color and depth information into one feature representation. Jia et al.~\cite{jia2014latent} proposed one transfer learning method that transferred the knowledge from depth information to the RGB dataset for effective RGB-based action recognition. Hu et al.~\cite{hu2015jointly} proposed a multi-task learning method to simultaneously explore the shared and
feature-specific components for heterogeneous features fusion. Sharing similar ideas, Kong and Fu~\cite{Kong2015CVPR} compressed and projected the
heterogeneous features to a shared space while Kong and Fu~\cite{kong2017max} learned both the shared space and independent private spaces to capture the useful information for action recognition. However, all these efforts are based on hand-crafted features and tend to be dataset-dependent. In this paper, we propose to encode the depth and RGB video into structured dynamic images, and exploit the conjoint information of the heterogeneous modalities using one c-ConvNet. This enhances the capability of the conventional ConvNet for action recognition from a single or multiple heterogeneous modalities.

\section{The Proposed Method}\label{hConvNets}
The proposed method consists of three phases, as illustrated in Figure~\ref{framework}, viz., the constructions of RGB visual dynamic images (VDIs) and depth dynamic images (DDIs), c-ConvNets and product-score fusion for final heterogeneous-feature-based action recognition. The first phase is an unsupervised learning process. It applies bidirectional rank pooling method to generate the VDIs and DDIs and represented by two dynamic images (forward (DDIf) and backward (DDIb)). In the following sections, we describe the three phases in detail. The rank pooling method~\cite{bilen2016dynamic}, that aggregates spatio-temporal-structural information from one video sequence into one dynamic image, is also briefly summarized.

\subsection{Construction of VDIs \& DDIs}

Rank pooling defines a rank function that encodes the video into one feature vector. Let the RGB/depth video sequence with $k$ frames be represented as  $<d_{1},d_{2},...,d_{t},...,d_{k}>$, where $d_{t}$ is the average of RGB/depth features over time up to $t$-frame or $t$-timestamp.  At each time $t$, a score $r_{t} = \omega^{T}\cdot d_{t}$ is assigned. The score satisfies $r_{i} > r_{j} \Longleftrightarrow i > j$. In general, more recent frames are associated with larger scores.  This process can be formulated as:

\begin{equation}
\begin{aligned}
\ \mathop{\arg\min}_{\omega} \dfrac{1}{2}\parallel \omega \parallel^{2} + \delta \sum\limits_{i > j} \xi_{ij}\\ s.t.~~ \omega^{T}\cdot(d_{i}-d_{j})\geq 1 -\xi_{ij}, \xi_{ij} \geq 0\\
\end{aligned},
\end{equation}

where $\xi_{ij}$ is the slack variable. Optimizing the above equation defines the rank function that maps a sequence of $k$ RGB/depth video frames to a single vector $\omega^{*}$. Since this vector aggregates information from all the frames in the sequence, it can be used as a video descriptor. The process of obtaining $\omega^{*}$ is called rank pooling. In this paper, rank pooling is directly applied on the pixels of RGB/depth frames and the $\omega^{*}$ is of the same size as RGB/depth frames and forms a dynamic image. Since in rank pooling the averaged feature up to time t is used to classify frame t, the pooled feature is biased towards beginning frames of the depth sequence, hence, frames at the beginning has more influence to $\omega^{*}$. This is not justifiable in action recognition as there is no prior knowledge on which frames are more important than other frames. Therefore, unlike the work of Bilen et al.~\cite{bilen2016dynamic}, the rank pooling is applied bidirectionally RGB/Depth sequences to reduce such bias.

\begin{figure}[t]
\begin{center}
{\includegraphics[height = 45mm, width = 85mm]{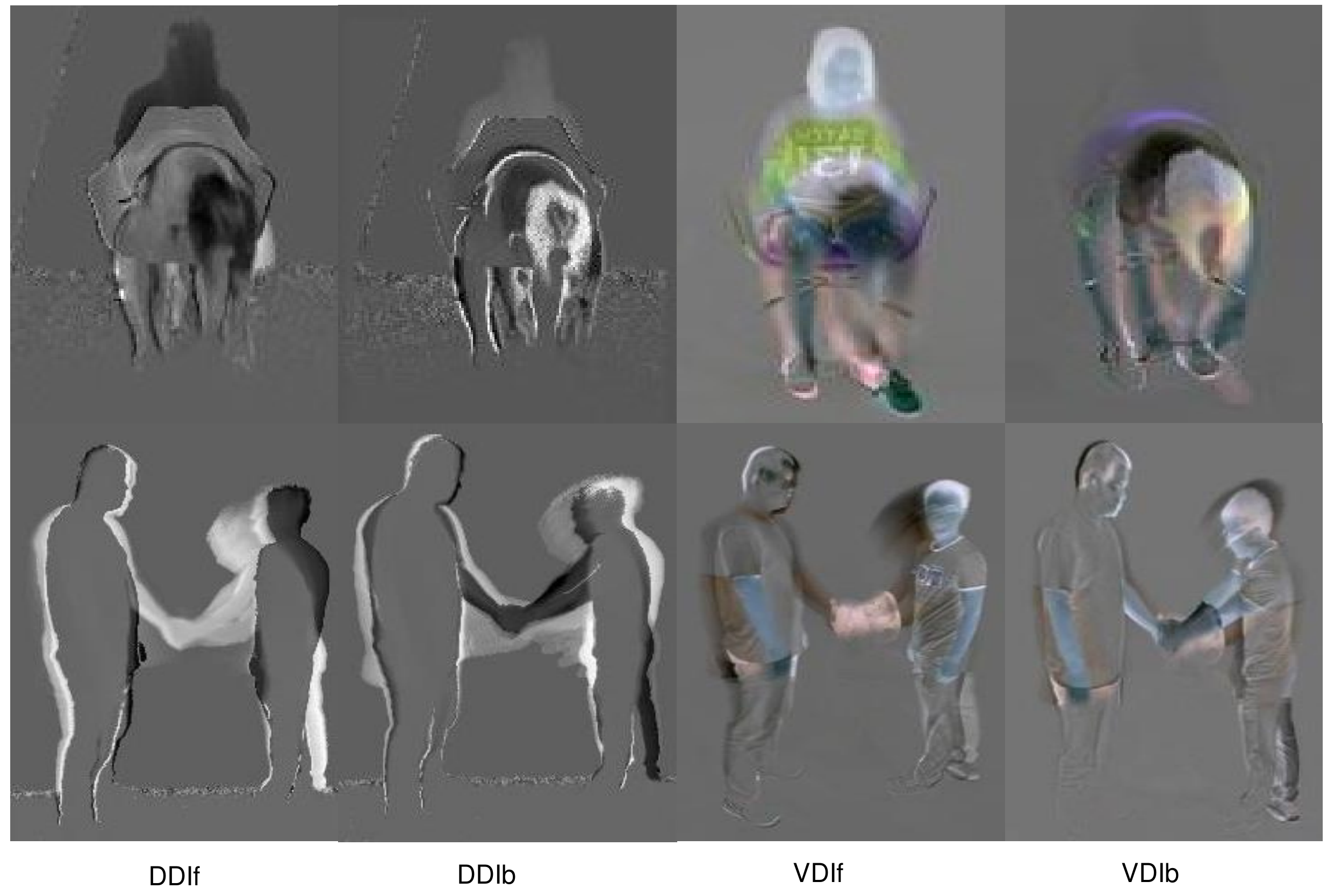}}
\end{center}
\caption{Visual comparisons of DDIf, DDIb, VDIf and VDIb.
The top row is the ``wear a shoe" action and the bottom row is
the action ``handshaking" from NTU RGB+D Dataset~\cite{shahroudy2016ntu}.   }
\label{DDIs}
\end{figure}

Visual comparisons of DDIf (forward), DDIb (backward), VDIf (forward) and VDIb (backward) are illustrated in Figure~\ref{DDIs}. From this figure, it can be seen that compared with VDIs,  DDIs lose the texture information of the object (shoes) and human, which is beneficial for simple action recognition without human-object interactions but not effective for interactions. The two directional DDIs and VDIs also capture different order of information for actions which are complementary to each other. Besides, the dynamic images also capture the structured information of an action, that  illustrates the coordination and synchronization of body parts over the period of the action, and describes the relations of spatial configurations of human body across different time slots.

\subsection{c-ConvNet}

\textbf{Joint Ranking and Classification}
The softmax loss adopted in the ConvNet can only learn separable features for homogeneous modalities, and is not guaranteed to be discriminative~\cite{wen2016discriminative}. In order to make the ConvNet more
discriminative for both RGB and depth modalities, the softmax and ranking losses are proposed to be jointly optimized as shown in
Figure~\ref{framework}. Triplet loss is a type of ranking loss, and has proven effective in several applications, such as face
recognition~\cite{schroff2015facenet,liu2016transferring}, pose estimation~\cite{kwak2016thin} and image retrieval~\cite{Jiang2016}. In this paper, the triplet loss is adopted as the ranking loss.  In common usage,
the triplet loss works on the homogeneous triplet data, namely, anchor, positive
and negative samples,  $(x_{a}^{i},x_{p}^{i},x_{n}^{i})$, where
$(x_{a}^{i},x_{p}^{i})$ have the same class label and $(x_{a}^{i},x_{n}^{i})$
have different class labels. The training encourages the network to find an
embedding $f(x)$ such that the distance between the positive sample and the anchor
sample $d_{<a,p>}^{i} = ||f(x_{a}^{i}) - f(x_{p}^{i})||_{2}^{2}$ is smaller than
the distance $d_{<a,n>}^{i} = ||f(x_{a}^{i}) - f(x_{n}^{i})||_{2}^{2}$ between
the negative sample and the anchor sample by a margin, $\alpha$.  Thus the triplet loss
$l$ can be formulated as:
\begin{equation}\label{1}
l = \sum_{i}^{N}[||f(x_{a}^{i}) - f(x_{p}^{i})||_{2}^{2} - ||f(x_{a}^{i}) - f(x_{n}^{i})||_{2}^{2} + \alpha]_{+},
\end{equation}
where $N$ is the number of possible triplets.

\begin{figure}[t]
\begin{center}
{\includegraphics[height = 50mm, width = 85mm]{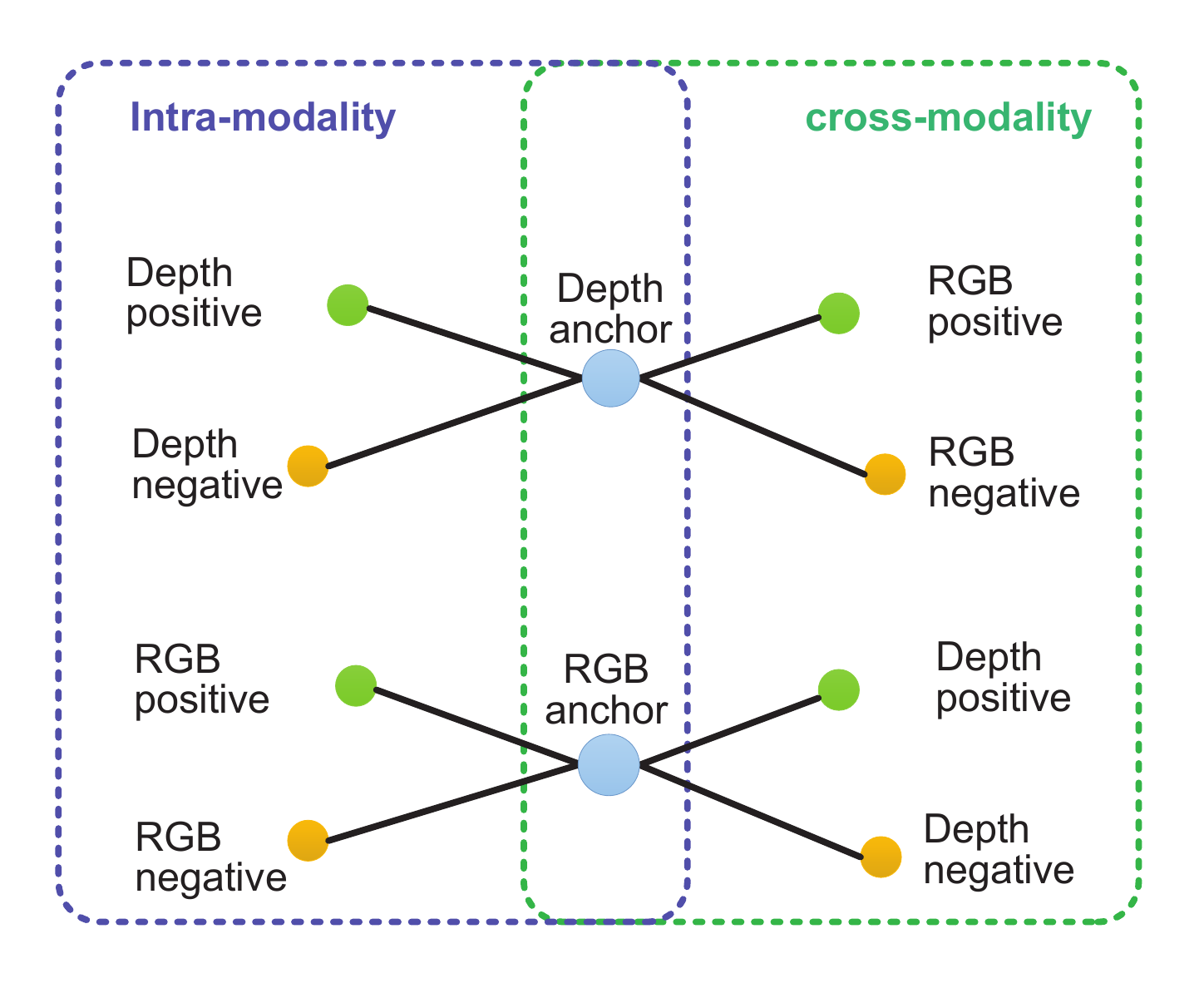}}
\end{center}
\caption{Illustration of the intra-modality and inter-modality triplets. }
\label{triplets}
\end{figure}

In order to make the triplet loss suitable for both homogeneous and heterogeneous modality-based recognition,
a new triplet loss consisting of both intra-modality and inter-modality triplets is designed (see Figure~\ref{triplets}). For the sake of computational efficiency and consideration of both intra and inter modalities
variations, four types of triplets are defined in this paper. If the anchor is one depth sample, then two positive and negative depth samples are assigned to intra-modality triplet while
two RGB samples are assigned to cross-modality triplet; if the anchor is one RGB
sample, then two positive and negative RGB samples are assigned to
intra-modality triplet while two depth samples are assigned to cross-modality
triplet.
Thus, the new ranking loss can be expressed as:
\begin{equation}\label{2}
L_{r} = (l^{Dep,Dep} + l^{RGB,RGB}) + \lambda(l^{Dep,RGB} + l^{RGB, Dep}),
\end{equation}
where $l^{Dep,Dep}$ denotes the intra-modality loss function of
triplet $(x_{a_{depth}}^{i},x_{p_{depth}}^{i},x_{n_{depth}}^{i})$; $l^{Dep,RGB}$
represents inter-modality loss function of triplet
$(x_{a_{depth}}^{i},x_{p_{RGB}}^{i},x_{n_{RGB}}^{i})$; and it is analogous
to $l^{RGB,RGB}$ and $l^{RGB, Dep}$; $\lambda$ trades off between the two kinds
of losses. With the constraint of these four triplet losses, the network is
forced more towards action distinction so that the cross-modality variance
is weakened or even eliminated. In this way, the knowledge about the
correlations between RGB and depth data are also incorporated in the c-ConvNet,
and enables the use of additional depth information for the case where only RGB
information is available.

Together with the softmax loss, the final loss function to be optimized in this paper is formulated as:
\begin{equation}\label{3}
L = L_{s} + \gamma L_{r},
\end{equation}
where $L_{s}$ denotes the softmax loss and $\gamma$ is a weight to balance the different loss functions.

\textbf{Network Structure}
The c-ConvNet consists of one feature extraction network, a branch each for
ranking loss and softmax loss, as illustrated in
Figure~\ref{framework}.
The feature extraction network is shared by the two losses and it can be any
available pre-trained network over ImageNet. In this paper,
VGG-16~\cite{simonyan2014very} network is adopted due to its promising results
in various vision tasks. The softmax loss branch is built on the FC8 layer which
is same as VGG-16. The ranking loss branch consists of one feature mapping
layer (FC9-128), one L2 normalization layer, one triplet selection layer and one
ranking loss layer. The feature mapping layer built on the FC7 layer of
VGG-16, aims to learn a compact representation for the triplet embedding.
Inspired by~\cite{schroff2015facenet}, L2 normalization layer is followed to
constrain the embedding to live on the hypersphere space. Triplets are selected
online using one triplet selection layer to generate the four kinds of triplets.
In this layer, every training sample will be selected as the anchor sample, and
its corresponding positive and negative samples randomly selected according to
Figure~\ref{triplets}. The ranking loss is built on the triplet selection layer
to minimize the loss according to Equation~\ref{2}. In order to leverage the
bidirectional information of videos, two c-ConvNets are trained separately
based on forward and backward dynamic images. An effective product-score fusion
method is adopted for final action recognition based on FC8 layer.

\subsection{Product-score Fusion}
Given a test RGB and depth video sequences, two pairs of dynamic
images, VDIf \& DDIf, and VDIb \& VDIb are constructed and fed into two
different trained c-ConvNets. For each image pair, product-score fusion is
used. The score vectors output of the weight sharing c-ConvNets are
multiplied in an element-wise manner, and then the resultant score vectors
(product-score) are normalized using $L_{1}$ norm. The two normalized score
vectors are multiplied, element-wise, and the max score in the resultant vector
is assigned as the probability of the test sequences. The index of this max
score corresponds to the recognized class label.

\begin{table}[!ht]
\centering
\caption{Results and comparison on the ChaLearn LAP IsoGD Dataset using ConvNet
and c-ConvNet. \label{ISOcom}}
\scalebox{0.8}{
\begin{tabular}{|c|c|}
\hline
Method & Accuracy \\
\hline
DDIf (ConvNet)  & 36.13\%\\
\hline
VDIf (ConvNet)  & 16.20\%\\
\hline
DDIb (ConvNet)  & 30.45\%\\
\hline
VDIb (ConvNet)  & 14.99\%\\
\hline
DDIf + VDIf (ConvNet)  & 33.64\%\\
\hline
DDIb + VDIb (ConvNet)  & 30.48\%\\
\hline
DDIf + DDIb (ConvNet)  & \textbf{37.52\%}\\
\hline
VDIf + VDIb (ConvNet)  & 17.60\%\\
\hline
DDIf + VDIf + DDIb + VDIb (ConvNet)  & 35.65\%\\
\hline
\hline
DDIf (c-ConvNet)  & 36.36\%\\
\hline
VDIf (c-ConvNet)  & 28.44\%\\
\hline
DDIb (c-ConvNet)  & 36.55\%\\
\hline
VDIb (c-ConvNet)  & 31.95\%\\
\hline
DDIf + VDIf (c-ConvNet)  & 41.01\%\\
\hline
DDIb + VDIb (c-ConvNet)  & 40.78\%\\
\hline
DDIf + DDIb (c-ConvNet)  & 40.08\%\\
\hline
VDIf + VDIb (c-ConvNet)  & 36.60\%\\
\hline
DDIf + VDIf + DDIb + VDIb (c-ConvNet)  & \textbf{44.80\%}\\
\hline
\end{tabular}}
\end{table}

\section{Experiments}\label{results}
The proposed method was evaluated on three benchmark RGB-D
datasets,
namely, two large ones, ChaLearn LAP IsoGD~\cite{wanchalearn} and NTU
RGB+D~\cite{shahroudy2016ntu} datasets, and a small one, SYSU 3D
HOI~\cite{hu2015jointly} dataset. These three datasets cover a wide range of
different types of actions including gestures, simple actions, daily activities,
human-object interactions and human-human interactions. In the following, we
proceed by briefly describing the implementation details and then present the
experiments and results.

\subsection{Implementation Details}
The proposed method was implemented using the Caffe
framework~\cite{jia2014caffe}. First, the feature extraction network was
fine-tuned on both depth and RGB modalities. Then, the c-ConvNet was trained 30
epochs. The initial learning rate was set to 0.001 and decreased by a factor of
10 every 12 epochs. The batch size was set as 50 images, with 5 actions in each
batch. The network weights are learned using the mini-batch stochastic gradient
descent with the momentum set to the value 0.9 and weight decay set to the value
0.0005. The parameter $\gamma$ was assigned the value 10 in order to ensure
that the two losses are of comparable magnitude. Parameters $\alpha$ and
$\lambda$ were assigned values that depend on the level of difficulty of the
datasets.

\subsection{ChaLearn LAP IsoGD Dataset} The ChaLearn LAP IsoGD
Dataset~\cite{wanchalearn} includes
47933 RGB-D depth sequences, with 249 gestures performed by 21 different individuals. The dataset is divided into training, validation and test sets. All three sets consist of
samples of different subjects to ensure that the gestures of one subject in the
validation and test sets will not appear in the training set. As the test set is
not available for public usage, we report the results on the validation set. For
this dataset, the margin $\alpha$ was set to 0.2. The parameter, $\lambda$, was
set to a value of 5 to solve the more difficult task of learning large
cross-modality discrepancy.

\textbf{Results.} To compare the ConvNet with the c-ConvNet, four ConvNets
(VGG-16) on DDIf, VDIf, DDIb and VDIb were trained separately for 40 epochs,
initialized with the pre-trained models over ImageNet. The initial learning rate
was set to 0.001 and decreased by a factor of 10 every 16 epochs.  The momentum
and weight decay parameters were set similarly as c-ConvNet. It is
found that 40 epochs were enough to achieve good results; increasing the
training epochs would not increase but even decreased the results.  For
c-ConvNet, two c-ConvNets are trained separately based on DDIf\&VDIf, and
DDIb\&VDIb, as illustrated in Figure~\ref{framework}. The trained c-ConvNet can
be used for single or both modalities testing. For
both cases, the product-score fusion method was adopted to aggregate different
channels.
The comparisons of ConvNet and c-ConvNet are shown in Table~\ref{ISOcom}.
From this Table it can be seen that for depth channels, DDIf and DDIb, the
c-ConvNet only increases the accuracy slightly, but for RGB
channels, VDIf and VDIb, the improvements are over 10 percentage points.
Interestingly, for ConvNet, due to the poor results of RGB features, the fusion
of additional RGB channels decreased the final accuracy compared with
those in which  only depth  was adopted. Meanwhile, the proposed c-ConvNet
significantly improved the RGB channel, and the fusion of two modalities
improved the final results.
These results demonstrate that knowledge about the correlations between
RGB and depth data are incorporated in the c-ConvNet, and enables the use of
additional depth information for the case where only RGB information is
available for testing. The fusion of both forward and backward dynamic images
improved the final accuracy by around 5 percentage points. Thus justifying
that bidirectional motion information are mutually beneficial and can
improve action recognition. The results of c-ConvNet in the final fusion
over the four channels improved by nearly 10 percentage points; a strong
demonstration of the effectiveness of the proposed method.

Table~\ref{table1} shows the comparisons of proposed method with previous works.
 Previous methods include MFSK combined 3D SMoSIFT~\cite{wan20143d} with
(HOG, HOF and MBH)~\cite{wang2013action} descriptors.  MFSK+DeepID further
included Deep hidden IDentity (Deep ID) feature~\cite{sun2014deep}. Thus, these
two methods utilized not only hand-crafted features but also deep learning
features. Moreover, they extracted features from RGB and depth separately,
concatenated them together, and adopted Bag-of-Words (BoW) model as the final
video representation. The other methods,
WHDMM+SDI~\cite{pichaoTHMS,bilen2016dynamic}, extracted features and conducted
classification with ConvNets from depth and RGB individually and adopted
product-score fusion for final recognition. SFAM~\cite{Pichaocvpr2017} adopted
scene flow to extract features and encoded the flow vectors into action maps,
which fused RGB and depth data from the onset of the process.
From this table, we can see that the proposed method outperformed all of these
recent works significantly, and illustrated its effectiveness.


\begin{table}[!ht]
\caption{Results and comparison on the ChaLearn LAP IsoGD Dataset with previous papers (D denotes Depth).  \label{table1}}
\centering
\scalebox{0.8}{
\begin{tabular}{|c|c|c|}
\hline
Method & Modality &Accuracy \\
\hline
MFSK~\cite{wanchalearn}  & RGB+D &18.65\%\\
\hline
MFSK+DeepID~\cite{wanchalearn} & RGB+D & 18.23\%\\
\hline
SDI~\cite{bilen2016dynamic} & RGB & 20.83\%\\
\hline
WHDMM~\cite{pichaoTHMS} & D & 25.10\%\\
\hline
WHDMM+SDI & RGB+D & 25.52\%\\
\hline
SFAM~\cite{Pichaocvpr2017} & RGB+D & 36.27\% \\
\hline
Proposed Method  & RGB+D & \textbf{44.80\%} \\
\hline
\end{tabular}}
\end{table}

 \begin{table*}[htbp]\small
\setlength{\belowcaptionskip}{9pt}
\caption{Comparative accuracies of the proposed method and previous methods on NTU RGB+D Dataset.\label{tableNTU} }
  \centering
\scalebox{1.0}{
 \begin{tabular}{|c|c|c|c|}
  \hline
  Method& Modality & CS	&  CV\\
    \hline
Lie Group~\cite{vemulapalli2014human}	& Skeleton &50.08\%	 & 52.76\%\\

HBRNN~\cite{du2015hierarchical} & Skeleton	&59.07\%	 & 63.97\%\\

2 Layer RNN~\cite{shahroudy2016ntu}	& Skeleton &56.29\% &	64.09\%\\

2 Layer LSTM~\cite{shahroudy2016ntu} & Skeleton	&60.69\%	&67.29\%\\

Part-aware LSTM~\cite{shahroudy2016ntu} & Skeleton &	62.93\%	&70.27\%\\

ST-LSTM~\cite{liu2016spatio} & Skeleton &65.20\%	&76.10\%	\\
Trust Gate~\cite{liu2016spatio} & Skeleton &	69.20\%	&77.70\%\\
\hline
HON4D~\cite{Oreifej2013} & Depth & 30.56\% & 7.26\%\\
SNV~\cite{yangsuper} & Depth & 31.82\% & 13.61\%\\
SLTEP~\cite{ji2017spatial} & Depth & 58 .22\% & -- \\
\hline
SSSCA-SSLM~\cite{shahroudy2017deep} & RGB+Depth & 74.86\% & --\\
Proposed Method	& RGB+Depth &\bfseries{86.42}\%&	\bfseries{89.08}\%\\
\hline

\end{tabular}}
\end{table*}

\subsection{ NTU RGB+D Dataset}
 To our best knowledge, NTU RGB+D Dataset is currently the largest action
recognition dataset in terms of training samples for each action. The 3D data is
captured by Kinect v2 cameras. The dataset has more than 56 thousand sequences
and 4 million frames, containing 60 actions performed by 40 subjects aged
between 10 and 35. It consists of front view, two side views and left, right 45
degree views. This dataset is challenging due to large intra-class and
viewpoint variations. For fair comparison and evaluation, the same protocol as
that in~\cite{shahroudy2016ntu} was used. It has both cross-subject and
cross-view evaluation. In the cross-subject evaluation, samples of subjects 1,
2, 4, 5, 8,
9, 13, 14, 15, 16, 17, 18, 19, 25, 27, 28, 31, 34, 35 and 38 were used as
training and samples of the remaining subjects were reserved for testing. In the
cross-view evaluation, samples taken by cameras 2 and 3 were used as training,
while the testing set includes samples from camera 1. For this dataset, the
margin $\alpha$ was set to 0.1 while $\lambda$ was set to 2.

\textbf{Results.} Similarly to LAP IsoGD Dataset, we conducted several
experiments to compare the conventional ConvNet and c-ConvNet, and the
comparisons are shown in Table~\ref{tableNTUcom}. From this table, we can
see that the c-ConNet learned more discriminative features compared
to conventional ConvNet.
Analysis of this results and the comparative results on LAP IsoGD Dataset
indicates that the improvements gained on NTU RGB+D Dataset are less than those
of LAP IsoGD Dataset. This is probably due to the high accuracy already
achieved on this dataset by ConvNet. From these two comparisons it may
be conclude that c-ConvNet works better on the difficult datasets for
recognition.

 \begin{table}[h]\small
\setlength{\belowcaptionskip}{9pt}
\caption{Results and comparison on the NTU RGB+D Dataset using ConvNet and
c-ConvNet.\label{tableNTUcom} }
  \centering
  \scalebox{0.85}{
\begin{tabular}{|c|c|c|}
  \hline
  Method&	Cross subject	&  Cross view\\
    \hline
DDIf (ConvNet)	&75.80\%	 & 76.50\%\\

VDIf (ConvNet)	&70.99\%	 & 75.45\%\\

DDIb (ConvNet)	&76.44\% &	75.62\%\\

VDIb (ConvNet)	&71.37\%	&76.57\%\\

DDIf + VDIf (ConvNet) &	80.77\%	&83.19\%\\

DDIb + VDIb (ConvNet) & 80.74\%	&83.04\%	\\

DDIf + DDIb (ConvNet) &	81.66\%	&81.53\%\\

VDIf + VDIb (ConvNet) & 78.31\%	& 83.58\%	\\
\tabincell{c}{DDIf + VDIf + \\
DDIb + VDIb (ConvNet)} &	\textbf{84.99\%}	&\textbf{87.51\%}\\
\hline
DDIf (c-ConvNet)	&76.58\%	 & 78.22\%\\

VDIf (c-ConvNet)	&71.35\%	 & 77.41\%\\

DDIb (c-ConvNet)	&77.69\% &	76.55\%\\

VDIb (c-ConvNet)	&73.24\%	&78.02\%\\

DDIf + VDIf (c-ConvNet) &	82.64\%	&85.21\%\\

DDIb + VDIb (c-ConvNet) &82.81\%	&85.62\%	\\
DDIf + DDIb (c-ConvNet) & 82.51\%	& 83.26\%\\

VDIf + VDIb (c-ConvNet) & 78.59\%	& 84.68\%	\\
\tabincell{c}{DDIf + VDIf +\\
DDIb + VDIb (c-ConvNet)} &	\textbf{86.42\%}	&\textbf{89.08\%}\\
\hline
\end{tabular}}
\end{table}

Table~\ref{tableNTU} lists the performance of the proposed method and those
previous works.
The proposed method was compared with some skeleton-based methods,
depth-based methods and RGB+Depth based methods that are previously reported on this dataset. We can see that the
proposed method outperformed all the previous works significantly.
Curious observation of the results shown in Table~\ref{tableNTUcom}
and Table~\ref{tableNTU} indicates that when only one channel of the dynamic
images (e.g. DDIf or VDIf) is adopted, the proposed method still achieved the
best results. This is a strong demonstration of the effectiveness of dynamic
images using ConvNets.

\begin{table}[!ht]
\centering
\caption{Results and comparison on the SYSU 3D HOI Dataset using ConvNet and
c-ConvNet. \label{SYSUcom}}
\scalebox{0.85}{
\begin{tabular}{|c|c|}
\hline
Method & Accuracy \\
\hline
DDIf (ConvNet)  & 97.92\%\\
\hline
VDIf (ConvNet)  & 91.25\%\\
\hline
DDIb (ConvNet)  & 92.50\%\\
\hline
VDIb (ConvNet)  & 92.92\%\\
\hline
DDIf + VDIf (ConvNet)  & 97.08\%\\
\hline
DDIb + VDIb (ConvNet)  & 94.58\%\\
\hline
DDIf + DDIb (ConvNet)  & 97.92\%\\
\hline
VDIf + VDIb (ConvNet)  & 93.33\%\\
\hline
DDIf + VDIf + DDIb + VDIb (ConvNet)  & \textbf{97.92\%}\\
\hline
\hline
DDIf (c-ConvNet)  & 97.92\%\\
\hline
VDIf (c-ConvNet)  & 92.50\%\\
\hline
DDIb (c-ConvNet)  & 92.50\%\\
\hline
VDIb (c-ConvNet)  & 92.50\%\\
\hline
DDIf + VDIf (c-ConvNet)  & 97.08\%\\
\hline
DDIb + VDIb (c-ConvNet)  & 95.00\%\\
\hline
DDIf + DDIb (c-ConvNet)  & 97.92\%\\
\hline
VDIf + VDIb (c-ConvNet)  & 95.00\%\\
\hline
DDIf + VDIf + DDIb + VDIb (c-ConvNet)  & \textbf{98.33\%}\\
\hline
\end{tabular}}
\end{table}

\begin{table}[ht!]
\centering
\caption{Comparison of the proposed method with previous approaches on SYSU 3D HOI Dataset.\label{table5}}
\scalebox{0.9}{
\begin{tabular}{|c|c|c|}
\hline
Method & Modality &Accuracy \\
\hline
HON4D~\cite{Oreifej2013} & Depth& 79.22\%  \\
\hline
MTDA~\cite{zhang2011multi} & RGB+Depth & 84.21\%\\
\hline
JOULE-SVM~\cite{hu2015jointly}& RGB+Depth & 84.89\%  \\
\hline
Proposed Method & RGB+Depth &\textbf{98.33\%}  \\
\hline
\end{tabular}}
\end{table}


\begin{figure}[t]
\begin{center}
{\includegraphics[height = 40mm, width = 85mm]{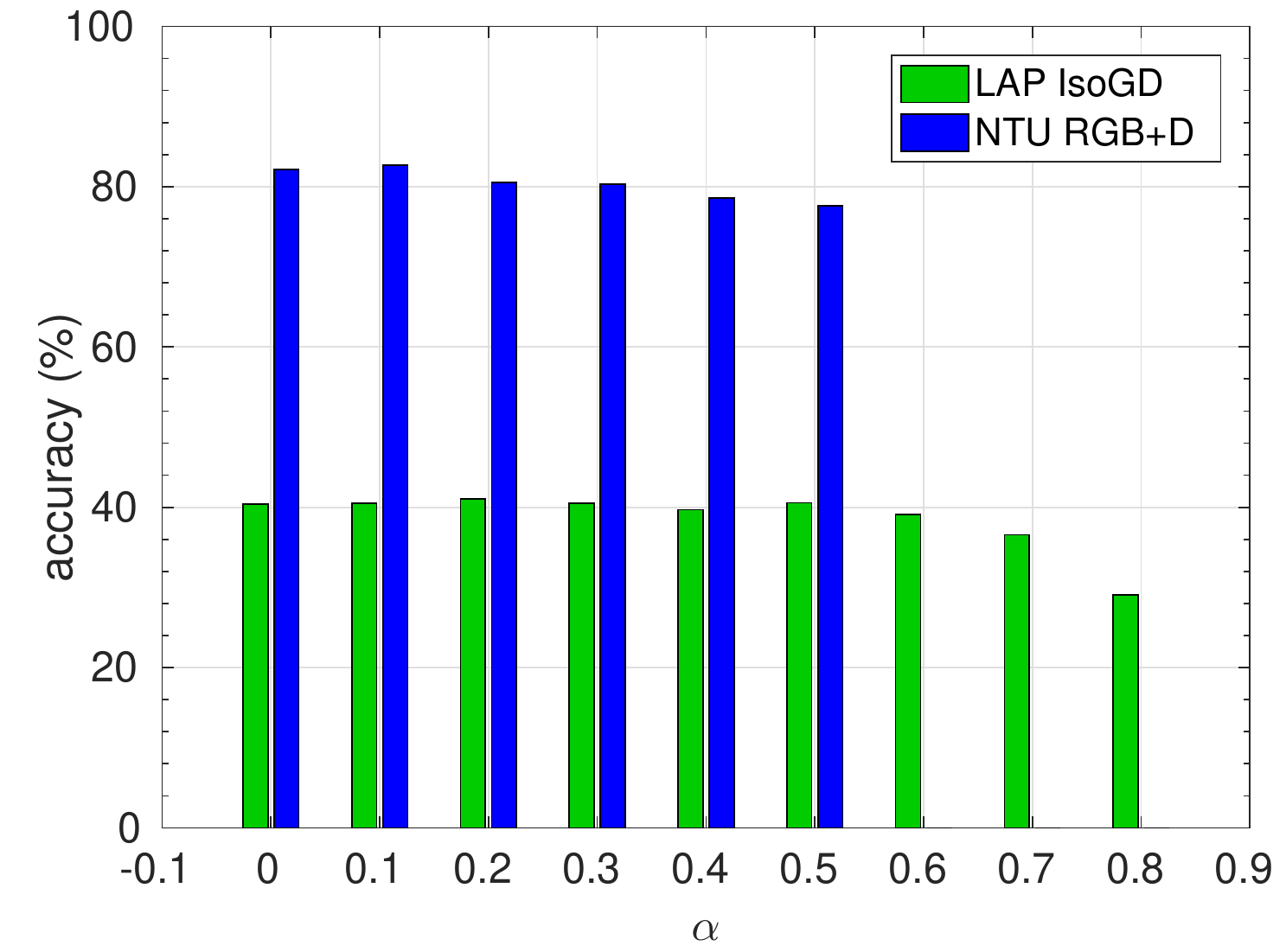}}
\end{center}
\caption{Comparison of margin $\alpha$ on LAP IsoGD and NTU RGB+D (Cross subject
setting) datasets in terms of accuracy(\%).\label{alpha} }
\end{figure}

\subsection{SYSU 3D HOI Dataset}
The SYSU 3D Human-Object Interaction Dataset (SYSU 3D HOI
Dataset)~\cite{hu2015jointly}  was collected to focus on human-object
interactions. There are 40 subjects performing 12 different activities. For each
activity, each participants manipulate one of the six different objects: phone,
chair, bag, wallet, mop and besom.
As this dataset is quite noisy, especially
the depth data, and the subjects are relatively small in the scene, the ranking pooling has been affected and the constructed DDIs and VDIs become noisy as well. Only 69\% recognition accuracy was achieved by using the noisy dynamic images. In order to reduce the noise impact,
skeleton data were used to locate the joints of subjects,
and around each joint ($16$ joints in total were selected for the body) one VDI or DDI was generated
and the VDIs or DDIs of all $16$ joints are stitched together into one VDI or DDI as input to the c-ConvNets.
For this dataset, the margin $\alpha$ was set to 0 while $\lambda$ was set to 1.

\textbf{Results.} Similarly to the above two large datasets, we conducted the
following experiments to compare the ConvNet and c-ConvNet as in
Table~\ref{SYSUcom}. From this table, it can be inferred that the proposed
method would still work on these small simple datasets, albeit with a slight
increase the final accuracy. Table~\ref{table5} compares the performances of the proposed method and those
of existing methods on this dataset using cross-subject settings as
in~\cite{hu2015jointly}. It can bee seen that, the proposed method outperformed
previous methods significantly.

\begin{table}[!th]
\centering
\caption{Comparison of weight $\lambda$ on LAP IsoGD and NTU RGB+D (Cross
subject setting) datasets in terms of accuracy(\%).\label{lambda}}
\footnotesize
\scalebox{0.9}{
\begin{tabular}{|c|c|c|c|c|c|c|} \hline
\multirow{3}{*}{Dataset}
& \multicolumn{6}{c|}{$\lambda$}\\
 \cline{2-7}
 & 0 & 1  &  2 & 3  & 5 & 7  \\ \hline
  \tabincell{c}{LAP\\IsoGD} & 39.68 & 39.51 & 39.61 & 39.71 & 41.01  & 40.13  \\ \hline
  \tabincell{c}{NTU\\RGB+D} & 80.36 & 81.15 & 82.64 & 80.18 & 80.06 & 80.11 \\ \hline
\end{tabular}}
\end{table}


\subsection{Further Analysis}
\subsubsection{Score-fusion}
In this paper, an effective product-score fusion method was adopted to improve
the final accuracy on the four-channel dynamic images. The other two commonly
used late score fusion methods are average  and maximum score fusion. The
comparisons among the three late score fusion methods are shown in
Table~\ref{table4.3}. We can see that the product-score fusion method achieved
the best results on all the three datasets. This verifies that the four-channel
dynamic images, namely, DDIf, VDIf, DDIb and VDIb, provide
mutually complementary information.

\begin{table}[!th]
\centering
\caption{Comparison of three different late score fusion methods on the three datasets.\label{table4.3}}
\scalebox{0.9}{
\begin{tabular}{|c|c|c|c|} \hline
\multirow{3}{*}{Dataset}
& \multicolumn{3}{c|}{Score Fusion Method}\\
 \cline{2-4}
 & Max  & Average  & Product \\ \hline
  LAP IsoGD & 42.01\%  & 43.48\% & \textbf{44.80\%}\\ \hline
  \tabincell{c}{NTU RGB+D\\(Cross subject)} & 84.69\%  & 85.86\% & \textbf{86.42\%}\\ \hline
  \tabincell{c}{NTU RGB+D\\(Cross view)} & 87.01\%  & 87.98\% & \textbf{89.08\%}\\ \hline
  SYSU 3D HOI & 97.08\%  & 97.92\% & \textbf{98.33\%} \\ \hline
\end{tabular}}
\end{table}

\subsubsection{Margin parameter, $\alpha$}
In the triplet loss, the parameter $\alpha$ refers to the
margin between the anchor/positive and negative. A small alpha
value enforces less on the similarities between the anchor/positive and
negative, but results in faster convergence for the loss.
On the other hand, a large alpha value may lead to a network with good
performance, but slow convergence during training.
The channel DDIf\&VDIf was taken for example on both LAP IsoGD and NTU RGB+D
datasets (cross subject setting) to illustrate the effects of this parameter,
and the comparisons are illustrated in Figure~\ref{alpha}. From the table it can be
seen that on LAP IsoGD Dataset, it achieved best accuracy when $\alpha$ was set
to 0.2, and with the with the increase of the $\alpha$, the accuracy
decreased significantly. On NTU RGB+D Dataset, best accuracy was obtained when
$\alpha$ was set to 0.1, and decreased dramatically when $\alpha$ increased.
This evidence suggests that the accuracy is sensitive to this
parameter, and it is advisable to set relatively small $\alpha$ values for
reasonable results.

\subsubsection{Weight parameter, $\lambda$}
In this section, the impact of the weight parameter, $\lambda$, as it
balances the intra-modality and inter-modality triplet losses is discussed. The
channel  DDIf\&VDIf were taken for example, and  the comparisons are shown in
Table~\ref{lambda}. From this Table, it can be seen that assigning a relatively
large weight $\lambda$ (i.e. putting more weight on cross-modality
triplet loss), will improve the final accuracy for the difficult datasets
(e.g. LAP IsoGD Dataset). However, the accuracy is comparatively less
sensitive to this parameter than $\alpha$.

\section{Conclusion}\label{conclusion}
In this paper, a novel c-ConvNet for RGB-D based action recognition was
proposed. It cooperatively exploits the information in RGB visual features
(VDI) and depth features (DDI) by
jointly optimizing a ranking loss and a softmax loss. The c-ConvNet enhances
the discriminative power of the deeply learned features and weakens the
modality discrepancy. Further, it can be used for both homogeneous and
heterogeneous modality-based action recognition. The ranking loss consists of
intra-modality and cross-modality triplet losses, and it reduces both the
intra-modality and cross-modality feature variations. State-of-the-art results
on three datasets demonstrated and justified the effectiveness of the proposed
method.

\section*{Acknowledgment}
This work was partially supported by the National Key Research and Development Plan (Grant No. 2016YFC0801002), the Chinese National Natural Science Foundation Projects $\sharp$61502491, $\sharp$61473291, $\sharp$61572501, $\sharp$61572536, $\sharp$61673052, $\sharp$61773392, $\sharp$61403405, Science and Technology Development Fund of Macau (No. 112/2014/A3). We gratefully acknowledge the support of NVIDIA Corporation with the donation of the Titan Xp GPU used for this research.
\small

\bibliographystyle{aaai}

\end{document}